\documentclass[runningheads]{llncs}
\usepackage[T1]{fontenc}
\usepackage{graphicx}
\usepackage{amsmath}
\usepackage{amssymb}
\usepackage{booktabs}
\usepackage{multirow}
\usepackage{algorithm}
\usepackage{algorithmic}
\usepackage[table]{xcolor}
\usepackage{hyperref}
\usepackage{pgfplots}
\usepgfplotslibrary{groupplots}
\pgfplotsset{compat=1.18}

\definecolor{shotblue}{HTML}{1F77B4}
\definecolor{shotorange}{HTML}{FF7F0E}
\definecolor{shotgreen}{HTML}{2CA02C}
\definecolor{shotred}{HTML}{D62728}

\begin{document}

\title{Where to Place the Query? Unveiling and Mitigating Positional Bias in In-Context Learning for Diffusion LLMs via Decoding Dynamics}
\titlerunning{Query Placement in dLLM ICL}
\author{Zhengheng Li\thanks{Corresponding author: \email{213220730@seu.edu.cn}}, Panrui Li, Xuyang Liu, Puzhi Xia}
\authorrunning{Z. Li et al.}
\institute{Southeast University\\
\email{\{213220730, 213241582, 213233851, 213231900\}@seu.edu.cn}}

\maketitle

\begin{abstract}
While In-Context Learning (ICL) is extensively studied in Autoregressive (AR) LLMs, its mechanism within Diffusion Large Language Models (dLLMs) remains largely unexplored. Unlike AR models restricted by unidirectional causal masking, dLLMs intrinsically utilize bidirectional attention, offering extensive spatial flexibility for query placement. Unfortunately, current practices conventionally inherit AR-style trailing-query templates, often overlooking the structural paradigm shift. This paper presents a comprehensive analysis unveiling that query position is actually a first-order variable in dLLMs. Through empirical decoupling, we demonstrate that positional variance impacts generation quality on par with example semantic quality. Internally, this positional sensitivity stems from a spatial ``Recency Effect'' in attention flow and task-dependent shifts in decoding trajectories. To mitigate this instability without ground-truth labels, we reveal that traditional single-step confidence ($C_{decoded}$) fails in dLLMs. Instead, we propose Average Confidence ($\overline{C}$), a novel metric tracking the iterative decoding process. By establishing the foundational spatial ICL baselines, we introduce Auto-ICL, a training-free adaptive routing strategy that dynamically optimizes query placement, robustly approaching oracle performance across heterogeneous reasoning and perception tasks.

\end{abstract}

\section{Introduction}

\begin{figure}[t]
    \centering
    \includegraphics[width=\textwidth]{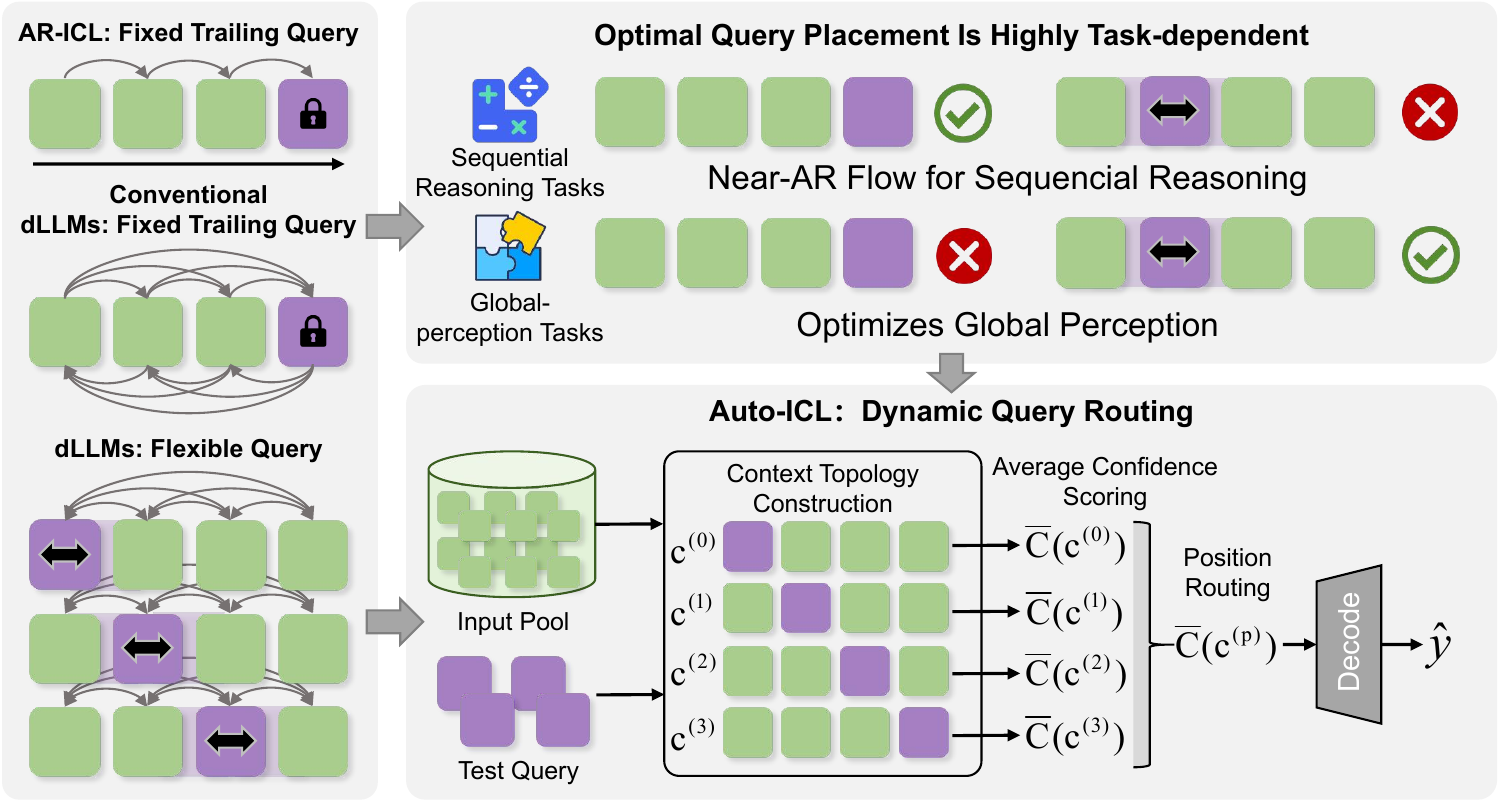}
    \caption{\textbf{Conceptual Overview of Positional Bias and Auto-ICL.} 
    \textbf{(Top) The Paradigm Shift and Positional Bias:} Unlike conventional AR-ICL which rigidly restricts the query to the trailing position, dLLMs unlock bidirectional spatial flexibility. However, optimal query placement is highly task-dependent: sequential reasoning tasks strictly favor trailing placement to maintain near-AR deduction flow, whereas global-perception tasks optimize better with prefix placement. 
    \textbf{(Bottom) Auto-ICL:} To dynamically identify the optimal topology without labels, our Auto-ICL framework enumerates candidate context topologies and scores them using trajectory-level Average Confidence ($\overline{C}$). This enables a training-free position router to execute the query at its task-specific optimal placement.}
    \label{fig:teaser}
\end{figure}

Diffusion Large Language Models (dLLMs), notably represented by recent models like LLaDA \cite{nie2025large} and Dream \cite{ye2025dream}, have emerged as a highly promising paradigm for sequence generation \cite{arriola2025block,wang2025diffusion,yu2025discrete}. By operating through an iterative decoding process, dLLMs intrinsically enable bidirectional context modeling \cite{wang2025time,wu2025fast,xu2025lopa,long2026focus}. This mechanism effectively circumvents the strict left-to-right causal constraints of conventional autoregressive (AR) models, facilitating a more comprehensive apprehension of global context and demonstrating significant potential in In-Context Learning (ICL) scenarios \cite{dong2024survey,workrethinking}.

However, a critical blind spot persists: while existing ICL literature predominantly investigates the selection \cite{liu2022makes,nguyen2023context,chang2023data,yoo2022ground,ghorbani2019data} and permutation \cite{lu2022fantastically,wu2023self,li2025order,bhope2025optiseq,guo2024makes} of demonstration examples, the spatial insertion position of the test query remains a largely unexplored variable. As conceptually illustrated in Figure~\ref{fig:teaser} (Top), in traditional AR models, the query is structurally rigid and inevitably restricted to the absolute end of the sequence due to inherent structural biases \cite{tang2023large,liu2024let} and recency anchoring \cite{zhao2021calibrate,cobbina2025show}. Conversely, the bidirectional attention mechanism in dLLMs grants substantial flexibility, allowing the query to be dynamically injected anywhere within the demonstration context. Unfortunately, current research conventionally applies AR-centric, trailing-query templates for dLLMs without reconsidering optimal placement \cite{li2025unveiling,li2026dip}.

To bridge this gap, this paper conducts a pioneering investigation into the positional sensitivity of ICL in dLLMs. Through rigorous empirical decoupling, we identify a fundamental property: query position is a first-order configuration variable. We demonstrate that the performance fluctuation induced solely by shifting the query position is on par with replacing the semantic content of the demonstrations themselves (yielding a variance impact ratio of $r=1.236$). Crucially, the optimal query placement is highly task-dependent: sequential reasoning tasks (e.g., GSM8K) strictly favor the traditional trailing placement, while global-perception tasks (e.g., Sudoku) achieve peak performance at the prefix boundary.

To demystify \textit{why} dLLMs exhibit such positional sensitivity, we probe the internal mechanistic dynamics across spatial and temporal dimensions. Spatially, via Attention Rollout \cite{abnar2020quantifying}, we identify a pronounced ``Recency Effect'' \cite{khandelwal2018sharp,zhang2025redundancy}, where the query disproportionately anchors its reasoning to physically adjacent examples. Temporally, query placement fundamentally reshapes the Decoding Trajectory \cite{li2025diffusion,huang2025pc}. A trailing query forces a near-AR, left-to-right generation pattern, whereas a prefix query triggers irregular, boundary-first decoding.

Recognizing that optimal query placement is crucial yet varies by task, we seek a method to dynamically route the query without ground-truth labels. We discover a fundamental metric mismatch: traditional single-step confidence ($C_{decoded}$), widely used in AR models, fails to rank spatial topologies in dLLMs because it discards the iterative temporal evolution. To resolve this, we propose Average Confidence ($\overline{C}$), a trajectory-level proxy that aggregates activation probabilities across all decoding steps, proving highly correlated with generation stability.

Since spatial query optimization in bidirectional dLLMs remains an uncharted territory, we formalize the foundational spatial configurations---Vanilla (trailing), Prefix, and Random---to serve as the first benchmarks for this new paradigm. Empowered by our trajectory-level metric, we design Auto-ICL (Figure~\ref{fig:teaser}, Bottom), a dynamic routing framework. Our core contributions are summarized as follows:
\begin{itemize}
    \item Empirical Decoupling of Positional Bias: We systematically quantify spatial query positional sensitivity in dLLMs, proving it is a first-order variable matching the impact of example selection.
    \item Spatiotemporal Mechanistic Insights: We trace the origins of positional bias to a spatial ``Recency Effect'' and temporal shifts in ``Decoding Trajectories,'' explaining the divergence in task-dependent optimal placements.
    \item Adaptive Position Routing (Auto-ICL): By replacing the flawed single-step confidence with trajectory-level Average Confidence ($\overline{C}$), we propose a fully unsupervised routing strategy. Auto-ICL dynamically identifies the most stable query placement, mitigating positional instability and approaching oracle performance across heterogeneous tasks.
\end{itemize}

% --- Section 2: RELATED WORK ---
\section{Related Work}
\label{sec:related_work}

To contextualize our contributions, we review recent advancements across three distinct but converging domains: the architectural evolution of diffusion language models, the configuration vulnerabilities of in-context learning, and the emerging exploration of ICL in bidirectional contexts.

\subsection{Diffusion Large Language Models (dLLMs)}
The dominance of Autoregressive (AR) sequence modeling has recently been challenged by the rapid advancement of Diffusion Large Language Models (dLLMs) \cite{arriola2025block,yu2025discrete}. By framing text generation as a discrete or continuous diffusion process \cite{wang2025diffusion}, dLLMs progressively decode a sequence from a fully masked state to clear text over multiple timesteps \cite{wang2025time,wu2025fast}. This iterative decoding paradigm inherently relies on bidirectional attention mechanisms, allowing the model to capture complex, non-causal global dependencies that strictly left-to-right AR models fail to apprehend \cite{xu2025lopa,long2026focus}. 

Furthermore, recent mechanistic interpretability studies indicate that dLLMs exhibit unique internal temporal dynamics. For instance, a recent research \cite{li2025diffusion} demonstrated that diffusion models implicitly form correct answers in their hidden states well before the tokens are explicitly decoded. Concurrently, sampling strategies such as PC-Sampler \cite{huang2025pc} highlight the importance of position-aware calibration in mitigating decoding biases. Despite these architectural and decoding breakthroughs, how to optimally elicit their reasoning capabilities via In-Context Learning remains severely underexplored.

\subsection{In-Context Learning and Prompt Configuration}
ICL has been proven to be remarkably powerful yet highly sensitive to prompt configuration \cite{dong2024survey}. In the context of traditional AR models, extensive research has established that ICL performance heavily fluctuates based on two primary factors: demonstration selection and demonstration ordering. Regarding selection, numerous studies have explored retrieving semantically similar examples \cite{liu2022makes}, utilizing influence functions \cite{nguyen2023context}, curating high-quality data pools \cite{chang2023data,ghorbani2019data}, and emphasizing the correctness of ground-truth labels \cite{yoo2022ground}. Regarding ordering, it is well-documented that arbitrary permutations of identical examples can cause accuracy to fluctuate significantly from state-of-the-art to near-random guessing \cite{lu2022fantastically,wu2023self,li2025order,bhope2025optiseq,guo2024makes}.

Beyond selection and ordering, AR models suffer from inherent structural biases \cite{tang2023large,liu2024let}. A prominent issue is \textit{recency bias}, where AR models overwhelmingly anchor their generation to the demonstrations placed nearest to the end of the sequence \cite{zhao2021calibrate,cobbina2025show}. However, all these prior works operate under a fundamental constraint: due to causal masking, the test query \textit{must} be fixed at the absolute end of the sequence. Consequently, the spatial configuration of the query itself has historically been treated as a rigid necessity rather than an optimizable variable.

\subsection{Bridging the Gap: ICL in Bidirectional Contexts}
With the advent of bidirectional dLLMs, the rigid causal constraint is lifted, unlocking substantial spatial flexibility for prompt construction. Recent preliminary efforts have begun exploring context configurations beyond standard AR constraints. For example, a recent work \cite{li2025unveiling} explored effective ICL configurations in multimodal tasks, while DIP \cite{li2026dip} proposed dynamic in-context planners for diffusion models. Moreover, insights from Vision-Language Models demonstrate that attention flow across reasoning tasks is highly redundant and position-dependent \cite{zhang2025redundancy,khandelwal2018sharp}, further emphasizing the need for spatial awareness \cite{abnar2020quantifying}.

However, when adapting NLP-based ICL to dLLMs, current practices often default to AR-style trailing-query templates, inadvertently suppressing the model's bidirectional potential. Our work bridges this critical gap. We move beyond simply reordering examples to explicitly optimizing the spatial topology of the query, treating \textit{Query Placement} as a first-order dimension and fundamentally adapting ICL to the spatiotemporal dynamics of dLLMs.

% --- Section 3: PROBLEM FORMULATION & SETUP ---
\section{Problem Formulation and Setup}

\subsection{ICL Formulation in dLLMs}
Given an ordered sequence of $N$ In-Context Examples (ICEs), denoted as $E = (e_1, e_2, \dots, e_N)$, each example represents an input-output pair $e_k = (x_k, y_k)$. Let $q$ be the test query containing $x_{\text{pred}}$. To systematically explore the impact of the query's spatial placement, we define $p \in \{0, 1, \dots, N\}$ as the insertion position index. The contextualized prompt $c^{(p)}$ is formulated as:
\begin{equation}
c^{(p)} = \left( \bigoplus_{k=1}^{p} e_k \right) \oplus q \oplus \left( \bigoplus_{k=p+1}^{N} e_k \right),
\label{eq:prompt}
\end{equation}
where $p=0$ indicates a prefix placement, $p=N$ indicates the standard trailing placement, and $0 < p < N$ corresponds to intermediate placements.

\subsection{Controlled Evaluation Protocol}
To isolate query position as the primary variable, we adopt a controlled protocol. For each task, we sample an ordered sequence of $N$ examples and keep both the semantic content and their internal ordering fixed. We then enumerate all insertion positions $p \in \{0, \dots, N\}$ for the test query. This ensures that performance variations are solely attributable to spatial positional differences rather than demonstration quality.

\subsection{Experiment Setup}
We utilize two representative open-source diffusion models: LLaDA-8B-Base and Dream-7B-Base, evaluated on two NVIDIA H100 GPUs. To systematically evaluate positional bias, we benchmark across five diverse datasets categorized into two distinct cognitive paradigms: \textbf{Sequential Reasoning Tasks} (GSM8K\cite{cobbe2021trainingverifierssolvemath} and MATH\cite{hendrycks2021measuringmassivemultitasklanguage} for mathematical deduction, alongside MBPP\cite{austin2021programsynthesislargelanguage} for code generation) and \textbf{Global-Perception Tasks} (Sudoku and Countdown\cite{ye2025autoregressiondiscretediffusioncomplex}). Unless stated otherwise, main experiments are conducted under a 4-shot setting.

% --- Section 4: EXTERNAL OBSERVATIONS ---
\section{External Sensitivity Experiments}

To assess the macro-level impact of query placement on ICL reasoning quality, we focus initially on Sudoku and GSM8K as representative benchmarks.

\subsection{Decoupling Query Position Sensitivity}
To investigate whether query placement introduces significant variance, we formulate an accuracy matrix $A \in \mathbb{R}^{M \times (N+1)}$, where $a_{m,p}$ denotes the accuracy using the $m$-th example set under the $p$-th query position. We define Query Position Sensitivity ($\overline{\sigma^{(M)}}$) and Example Set Sensitivity ($\overline{\sigma^{(P)}}$):
\begin{equation}
    \overline{\sigma^{(M)}} = \frac{1}{M} \sum_{m=1}^{M} \text{std}(a_{m,0}, a_{m,1}, \dots, a_{m,N}),
    \label{eq:query_sensitivity}
\end{equation}
\begin{equation}
    \overline{\sigma^{(P)}} = \frac{1}{N+1} \sum_{p=0}^{N} \text{std}(a_{1,p}, a_{2,p}, \dots, a_{M,p}),
    \label{eq:example_sensitivity}
\end{equation}
\begin{equation}
    r = \frac{\overline{\sigma^{(P)}}}{\overline{\sigma^{(M)}}},
    \label{eq:ratio}
\end{equation}

\begin{figure}[b!]
    \centering
    \includegraphics[width=\linewidth]{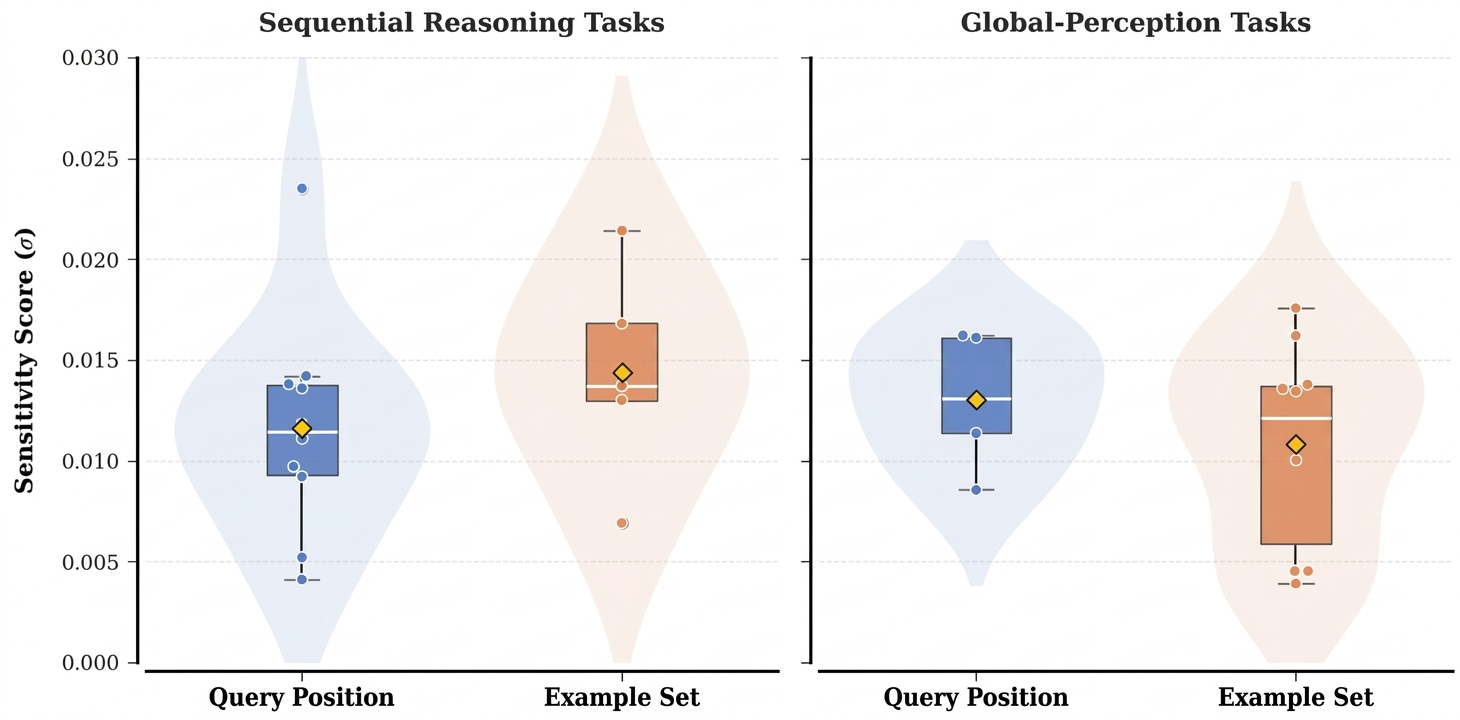}
    \caption{Violin plots illustrating the distributions of Query Position Sensitivity ($\overline{\sigma^{(M)}}$) and Example Set Selection Sensitivity ($\overline{\sigma^{(P)}}$). The red diamonds denote the mean standard deviations.}
    \label{fig:sensitivity_violin}
\end{figure}

\textbf{Finding 1: Query position is a first-order variable in dLLM in-context learning.} Position must be optimized jointly with example selection rather than treated as a fixed formatting choice. As shown in Table~\ref{tab:sensitivity_metrics} and Figure~\ref{fig:sensitivity_violin}, evaluating GSM8K yields an $r$ value of 1.236. An $r$ value approximating 1.0 indicates that variance triggered solely by shifting the query position is of a comparable magnitude to variance caused by entirely swapping the semantic content of the demonstrations. 

\begin{table}[htbp]
    \centering
    \caption{Sensitivity metrics showing Query Position Sensitivity, Example Set Sensitivity, and the Relative Importance Ratio across datasets.}
    \label{tab:sensitivity_metrics}
    \resizebox{\textwidth}{!}{%
    \begin{tabular}{lccc}
        \toprule
        \textbf{Dataset} & \textbf{Position Sensitivity ($\overline{\sigma^{(M)}}$)} & \textbf{Example Sensitivity ($\overline{\sigma^{(P)}}$)} & \textbf{Ratio ($r$)} \\
        \midrule
        Sudoku & 0.01723 & 0.01459 & 0.847 \\
        GSM8K & 0.01162 & 0.01437 & 1.236 \\
        Countdown & 0.02351 & 0.01368 & 0.582 \\
        MBPP & 0.01291 & 0.01398 & 1.083 \\
        MATH & 0.01274 & 0.01441 & 1.132 \\
        \bottomrule
    \end{tabular}%
    }
\end{table}

\subsection{Impact of Query Position on Performance}
\begin{figure}[htbp]
    \centering
    \includegraphics[width=\linewidth]{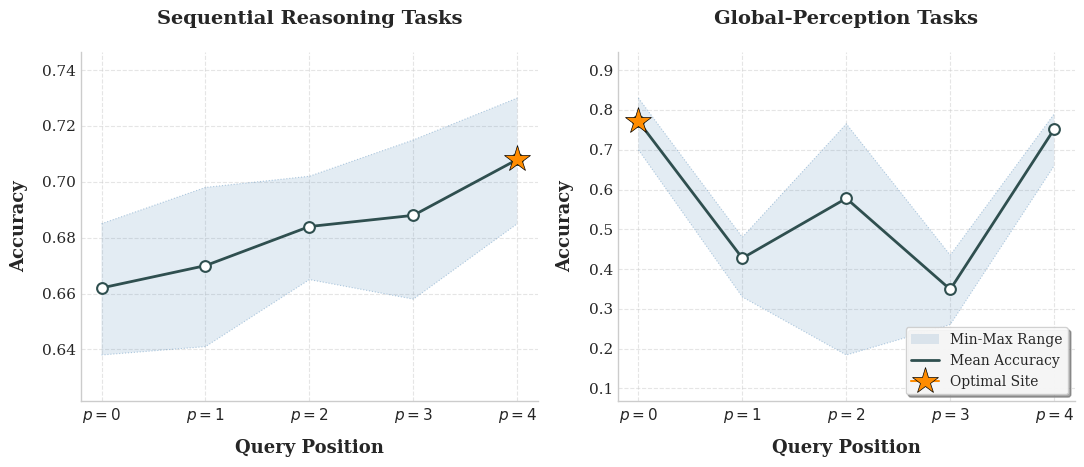}
    \caption{Accuracy as a function of query insertion position on GSM8K (left) and Sudoku (right). GSM8K exhibits \emph{trailing rigidity}, while Sudoku shows a \emph{non-monotone} profile peaking at the prefix position.}
    \label{fig:position_curves}
\end{figure}

\textbf{Finding 2: The optimal query position is highly task-dependent rather than universally trailing.} As illustrated in Figure~\ref{fig:position_curves}, for reasoning-intensive tasks (GSM8K), performance peaks when the query is at the trailing position. In contrast, tasks requiring global perception (Sudoku) achieve their best fixed-position accuracy at the prefix boundary.

% --- Section 5: INTERNAL MECHANISMS ---
\section{Internal Mechanism Analysis}
We demystify this positional bias by probing the dLLM's internal dynamics across spatial and temporal dimensions.

\subsection{Spatial Dimension: Recency Effect in Attention Flow}
To quantify information propagation, we introduce Attention Flow ($\mathcal{F}_k^{(p)}$), which aggregates attention weights from query tokens to the $k$-th ICE tokens across all decoding steps and layers:
\begin{equation}
\mathcal{F}_k^{(p)} = \frac{1}{|S_{Q^{(p)}, E_k}|} \sum_{(i,j) \in S_{Q^{(p)}, E_k}} \bar{A}(i, j),
\end{equation}

where $\bar{A}(i, j)$ denotes the averaged attention weight from query token $i$ to demonstration token $j$ across all layers and decoding steps, and $S_{Q^{(p)}, E_k}$ represents the set of all valid index pairs between the test query $Q^{(p)}$ and the $k$-th in-context example $E_k$.

\begin{figure}[htbp]
    \centering
    \includegraphics[width=0.85\linewidth]{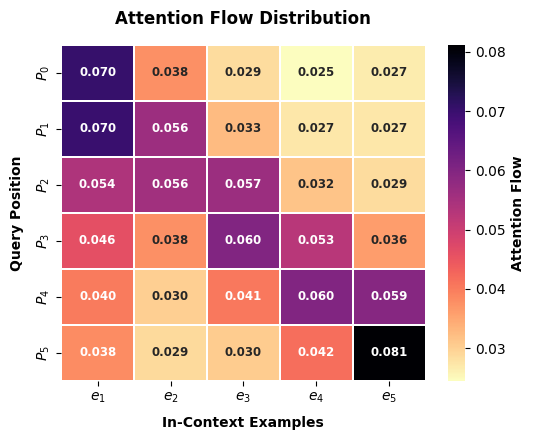}
    \caption{Attention-flow heatmap over query insertion boundaries and in-context examples on Sudoku. The diagonal concentration highlights a strong recency effect.}
    \label{fig:attention_flow}
\end{figure}

\textbf{Finding 3: Positional sensitivity is spatially governed by a Recency Effect.} Changing the query boundary alters which neighboring demonstrations dominate the reasoning context. As visualized in Figure~\ref{fig:attention_flow}, the strongest attention flow consistently concentrates on demonstrations adjacent to the query. While supplementary layer-wise analyses reveal that shallow layers may attend to uninformative ``anchor tokens'', the deep semantic layers strictly adhere to spatial locality.

\subsection{Temporal Dimension: Decoding Order}

\textbf{Finding 4: Positional sensitivity temporally arises from task-dependent decoding trajectories.} Strict left-to-right generation is crucial for c deduction (GSM8K), while non-linear trajectories help avoid local optima in global-pattern tasks (Sudoku). 

\begin{figure}[htbp]
    \centering
    \includegraphics[width=\textwidth]{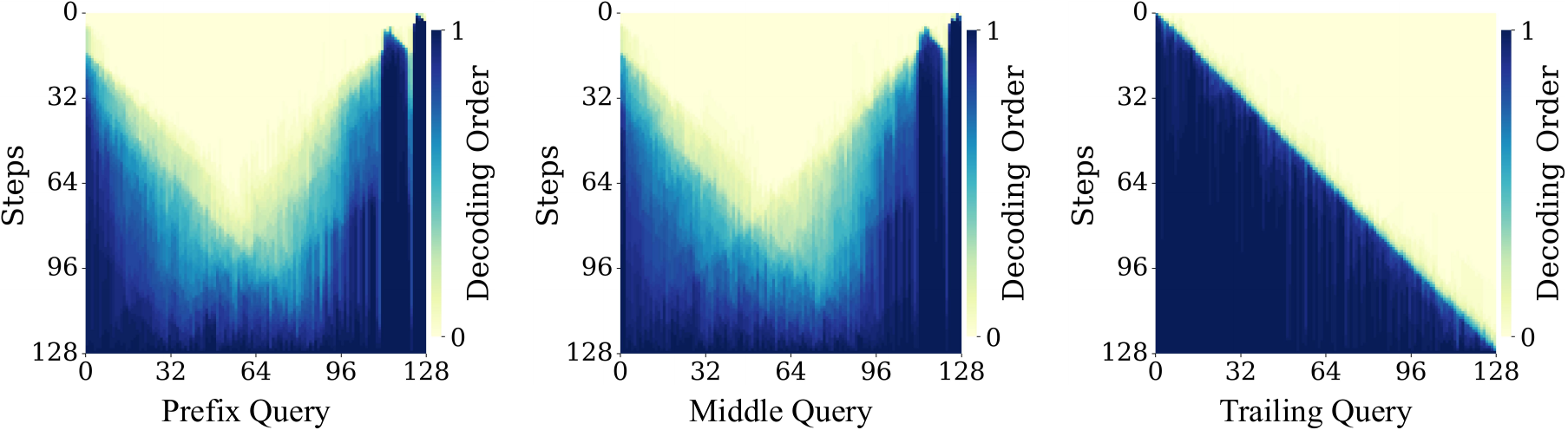}
    \caption{Average decoding-order heatmaps on 200 GSM8K instances across different query placements. \textbf{Trailing Query} actively enforces a strict left-to-right, AR-style diagonal trajectory, which is mandatory for sequential deduction. \textbf{Prefix Query} and \textbf{Middle Query} trigger a V-shaped, boundary-first decoding dynamic.}
    \label{fig:decoding_order}
\end{figure}

To uncover \textit{how} the spatial query placement fundamentally reshapes the temporal dynamics of dLLMs, we visualize the token-level decoding order across the diffusion timesteps. As shown in Figure~\ref{fig:decoding_order}, the x-axis corresponds to the generated sequence positions, the y-axis represents the iterative decoding steps (from $T=128$ down to $0$), and the color gradient indicates the chronological order of token crystallization.

We observe a striking morphological transformation driven solely by the query position:
\begin{itemize}
    \item \textbf{Trailing Query (Right Panel):} The decoding trajectory forms a strict, clean diagonal line from top-left to bottom-right. This indicates that placing the query at the end actively forces the dLLM to mimic the strict left-to-right causal generation of AR models. This sequential constraint is an absolute prerequisite for logic-heavy tasks like GSM8K, where each deductive step causally depends on the immediate predecessor.
    \item \textbf{Prefix Query (Left Panel):} The trajectory morphs into a profound ``V-shape'' pattern. Here, the model operates in a boundary-first manner, concurrently decoding both ends of the sequence before converging to fill in the middle. 
    \item \textbf{Middle Query (Center Panel):} The temporal dynamic collapses into a complex, multi-modal convergence pattern with competing local structures.
\end{itemize}
While these non-linear, bidirectional trajectories (Prefix/Middle) severely disrupt the causal chain required for sequential reasoning (explaining the GSM8K performance drop), they intrinsically provide the global topological apprehension necessary to satisfy the complex multi-way constraints of perception tasks like Sudoku.

\subsection{Average Confidence as a Probe for Accuracy}
Inspired by prior studies on confidence-based stopping signals in diffusion generation, we revisit the iterative decoding process of dLLMs. To quantify generation stability, we introduce Average Confidence ($\overline{C}$), a sequence-level metric that leverages temporal dynamics.

Formally, given a constructed context $c^{(p)}$, let $I$ denote the token index set of the target answer region, and $\mathcal{V}$ represent the model vocabulary. At decoding step $t$, the confidence score $s_{i}^{(t)}$ of token $i$ is the maximum activation probability: $s_{i}^{(t)} = \max_{v \in \mathcal{V}} P_{i,v}^{(t)}$. We define $\overline{C}$ as the expected mean of the confidence across all $T+1$ decoding steps:
\begin{equation}
\overline{C} = \frac{1}{|I| \cdot (T+1)} \sum_{t=0}^{T} \sum_{i \in I} s_{i}^{(t)},
\label{eq:avgconf}
\end{equation}
Thus, $\overline{C}$ serves as a trajectory-level proxy for position quality, summarizing the full decoding history.

% --- Section 6: APPLICATION ---
\section{Adaptive Position Routing (Auto-ICL)}

Our analyses in Sections 4 and 5 establish two critical premises. First, query placement is a first-order variable with highly task-dependent optima (Findings 1 and 2). Second, this performance variance is mechanistically driven by spatial recency effects and temporal shifts in the decoding trajectories (Findings 3 and 4). Because the optimal position oscillates based on the cognitive nature of the task, relying on a static AR-style trailing template is fundamentally inadequate for bidirectional dLLMs.

To unlock the full potential of bidirectional ICL, we must dynamically route the query to its optimal task-specific topology. Since ground-truth labels are unavailable at inference time, we require a robust proxy to evaluate the stability of different decoding trajectories. Drawing upon our mechanism analysis (Section 5.3), we leverage Average Confidence ($\overline{C}$) as this trajectory-level proxy. By holistically scoring the iterative decoding dynamics, we propose \textbf{Auto-ICL}, a training-free adaptive routing strategy that automatically identifies the optimal query placement.

\subsection{The Auto-ICL Algorithm}
\begin{figure}[htbp]
    \centering
    \includegraphics[width=\textwidth]{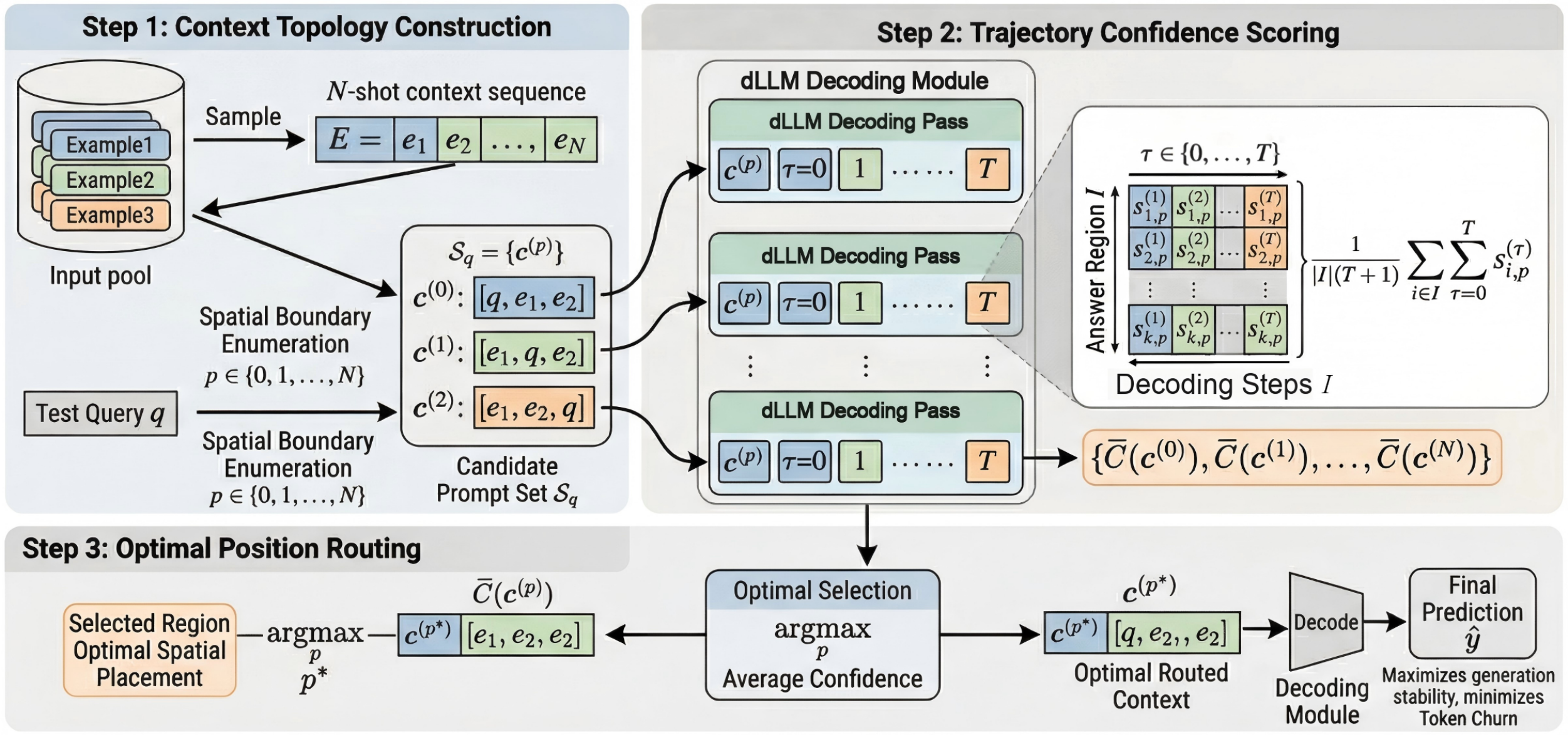}
    \caption{Overview of the Auto-ICL routing pipeline. For a given test query, we enumerate all valid spatial insertion boundaries, execute dLLM decoding passes to compute trajectory-level average confidence, and dynamically route the query to the position that maximizes generation stability.}
    \label{fig:auto_icl_pipeline}
\end{figure}

Given a candidate pool $\mathcal{D}$ and a test query $q$, the routing procedure consists of three steps, detailed in Algorithm~\ref{alg:routing} and Figure~\ref{fig:auto_icl_pipeline}.

\textbf{Step 1: Context Construction.} Sample $N$ ICEs to form sequence $E$. Construct the probing set $\mathcal{S}_{q} = \{c^{(p)} \mid p \in \{0, \dots, N\}\}$.

\textbf{Step 2: Confidence Scoring.} For each $c^{(p)}$, run a full dLLM decoding pass and compute $\overline{C}(c^{(p)})$ per Eq.~(\ref{eq:avgconf}).

\textbf{Step 3: Optimal Selection.} Return the placement that maximizes stability:
\begin{equation}
p^* = \arg\max_{p \in \{0,\dots,N\}} \overline{C}(c^{(p)}),
\label{eq:routing}
\end{equation}
The final answer is decoded from $c^{(p^*)}$. The method is entirely training-free and requires no labels at test time.

\begin{algorithm}[t]
\caption{Auto-ICL: Adaptive Position Routing via Average Confidence}
\label{alg:routing}
\begin{algorithmic}[1]
\REQUIRE Candidate pool $\mathcal{D}$, test query $q$, shot count $N$, decoding steps $T$
\ENSURE Optimal insertion position $p^*$ and final prediction $\hat{y}$
\STATE Sample ICE sequence $E = (e_1, \dots, e_N)$ from $\mathcal{D}$
\FOR{$p = 0, 1, \dots, N$}
    \STATE Construct context $c^{(p)}$ by inserting $q$ at position $p$
    \STATE Run dLLM decoding pass: obtain $\{s_{i,p}^{(t)}\}_{i \in I, t=0}^{T}$
    \STATE Compute $\overline{C}(c^{(p)}) \leftarrow \frac{1}{|I|(T+1)} \sum_{i \in I} \sum_{t=0}^{T} s_{i,p}^{(t)}$
\ENDFOR
\STATE $p^* \leftarrow \arg\max_{p}\,\overline{C}(c^{(p)})$
\STATE $\hat{y} \leftarrow \mathrm{Decode}(c^{(p^*)})$
\RETURN $p^*,\, \hat{y}$
\end{algorithmic}
\end{algorithm}

% --- Section 7: EVALUATION ---
\section{Evaluation and Ablations}

\subsection{Main Results}
Since spatial query routing in bidirectional dLLMs is an entirely novel problem setting, no existing routing methods are directly applicable. Therefore, we establish Vanilla (the AR-standard trailing placement, $p=N$), Prefix ($p=0$), and Random configurations as the foundational spatial baselines for this new paradigm. We evaluate Auto-ICL against these baselines, alongside an \textit{Oracle} configuration representing the theoretical upper bound. 

\begin{figure}[htbp]
    \centering
    \includegraphics[width=0.65\textwidth]{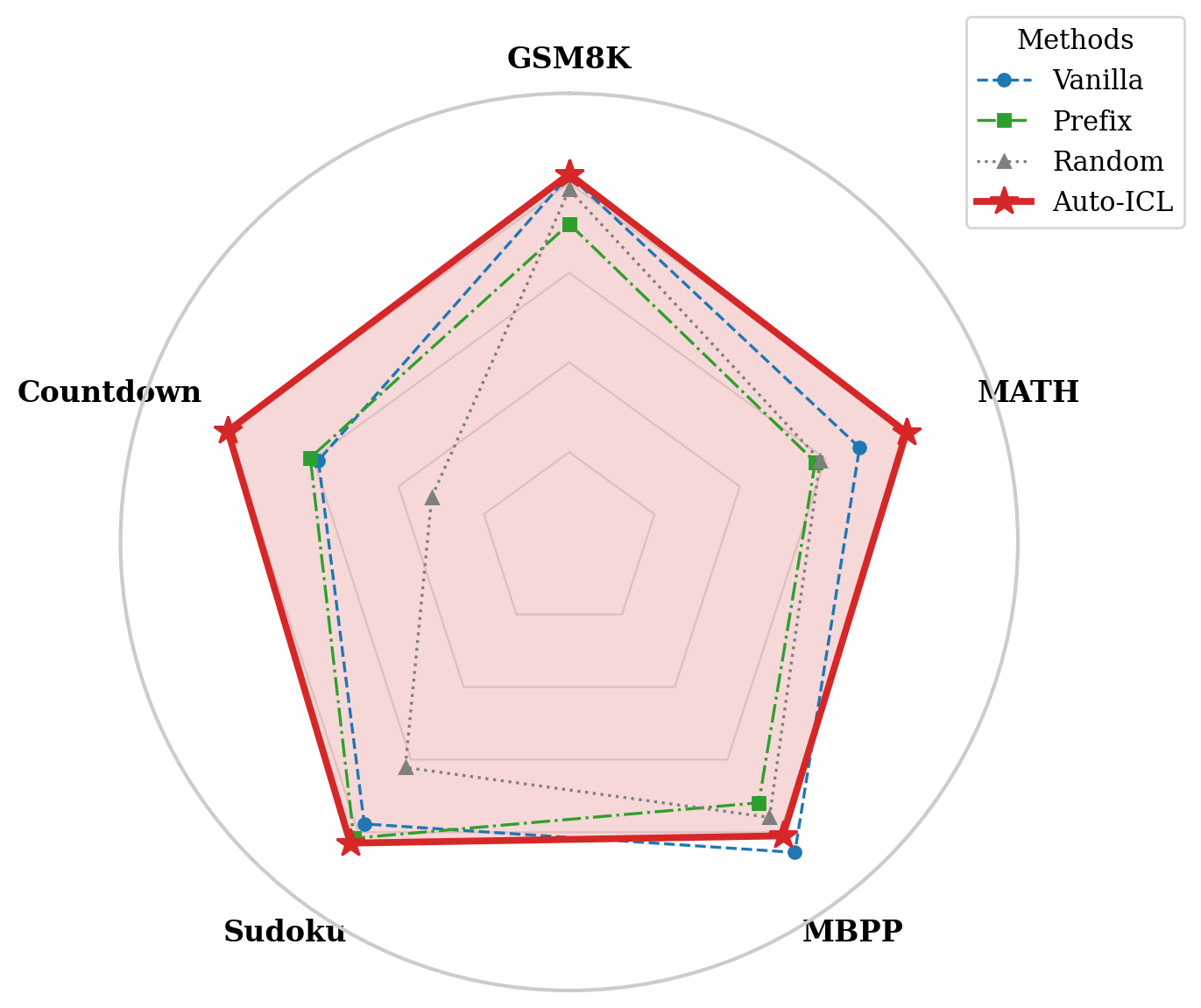}
    \caption{\textbf{Performance footprint across heterogeneous tasks.} The radar chart visualizes the accuracy distribution of various spatial routing strategies. Auto-ICL (red solid line) adaptively envelops the best-performing static baselines across all axes, effectively functioning as a universal upper-bound envelope for both sequential reasoning and global-perception tasks.}
    \label{fig:radar_chart}
\end{figure}

As detailed in Table~\ref{tab:main_results} and visually summarized in Figure~\ref{fig:radar_chart}, Auto-ICL robustly matches or surpasses the best available static placement across heterogeneous tasks. The radar chart (Figure~\ref{fig:radar_chart}) powerfully illustrates the dynamic adaptability of our method. While static strategies exhibit severe task-dependency—for instance, Vanilla excels on sequential tasks (GSM8K) but collapses on perception tasks (Sudoku)—Auto-ICL dynamically morphs to push the performance boundary on nearly every axis. Specifically, on Sudoku, it successfully recovers the non-trailing optimum (84.4\% on LLaDA); simultaneously on GSM8K, it strictly preserves the necessary sequential rigidity (71.0\%).

Crucially, this dynamic adaptability comes at almost no cost. As shown in Table~\ref{tab:main_results}, Auto-ICL achieves substantial accuracy gains with only a marginal increase in inference latency (e.g., +0.08s overhead on GSM8K). Because the candidate context topologies can be efficiently evaluated in parallel, the trajectory-level confidence routing introduces negligible time overhead compared to static configurations. By effectively forming an outer performance envelope across diverse cognitive dimensions without sacrificing efficiency, these results prove the universal efficacy and practicality of our Auto-ICL framework.

\begin{table}[htbp]
\centering
\caption{Accuracy and inference latency comparison of query placement strategies across datasets (4-shot). The proposed Auto-ICL is highlighted with a light blue background. \textit{Oracle} represents the theoretical upper bound utilizing ground-truth labels for routing, thus denoted in gray italics.}
% --- 尺寸控制区域开始 ---
\small
\setlength{\tabcolsep}{4pt}
\renewcommand{\arraystretch}{1.25}
% --- 尺寸控制区域结束 ---
\begin{tabular}{llcccc}
\toprule
\multirow{2}{*}{\textbf{Dataset}} & \multirow{2}{*}{\textbf{Method}} & \multicolumn{2}{c}{\textbf{LLaDA-8B-Base}} & \multicolumn{2}{c}{\textbf{Dream-7B-Base}} \\
\cmidrule(lr){3-4} \cmidrule(lr){5-6}
& & Acc (\%) $\uparrow$ & Latency (s) $\downarrow$ & Acc (\%) $\uparrow$ & Latency (s) $\downarrow$ \\
\midrule
\multirow{5}{*}{GSM8K}
& Vanilla         & 70.8 & 9.06 & \textbf{74.4} & 8.57 \\
& Prefix          & 61.4 & 9.09 & 63.3 & 8.59 \\
& Random          & 68.0 & 9.08 & 66.0 & 8.58 \\
& \cellcolor{blue!8}\textbf{Auto-ICL} & \cellcolor{blue!8}\textbf{71.0} & \cellcolor{blue!8}9.14 & \cellcolor{blue!8}71.2 & \cellcolor{blue!8}8.58 \\
& \textcolor{gray}{\textit{Oracle}} & \textcolor{gray}{\textit{82.8}} & \textcolor{gray}{\textit{N/A}} & \textcolor{gray}{\textit{81.0}} & \textcolor{gray}{\textit{N/A}} \\
\midrule
\multirow{5}{*}{MATH}
& Vanilla         & 23.6 & 23.17 & 30.0 & 18.16 \\
& Prefix          & 20.0 & 23.19 & 22.8 & 18.17 \\
& Random          & 20.4 & 23.17 & 23.1 & 18.15 \\
& \cellcolor{blue!8}\textbf{Auto-ICL} & \cellcolor{blue!8}\textbf{27.4} & \cellcolor{blue!8}23.24 & \cellcolor{blue!8}\textbf{32.2} & \cellcolor{blue!8}18.15 \\
& \textcolor{gray}{\textit{Oracle}} & \textcolor{gray}{\textit{36.6}} & \textcolor{gray}{\textit{N/A}} & \textcolor{gray}{\textit{39.2}} & \textcolor{gray}{\textit{N/A}} \\
\midrule
\multirow{5}{*}{MBPP}
& Vanilla         & \textbf{48.8} & 5.38 & \textbf{65.2} & 4.58 \\
& Prefix          & 41.0 & 5.35 & 46.3 & 4.60 \\
& Random          & 43.3 & 5.34 & 47.2 & 4.59 \\
& \cellcolor{blue!8}\textbf{Auto-ICL} & \cellcolor{blue!8}46.2 & \cellcolor{blue!8}5.40 & \cellcolor{blue!8}57.3 & \cellcolor{blue!8}4.60 \\
& \textcolor{gray}{\textit{Oracle}} & \textcolor{gray}{\textit{63.1}} & \textcolor{gray}{\textit{N/A}} & \textcolor{gray}{\textit{66.7}} & \textcolor{gray}{\textit{N/A}} \\
\midrule
\multirow{5}{*}{Sudoku}
& Vanilla         & 79.0 & 0.76 & 30.0 & 0.67 \\
& Prefix          & 83.0 & 0.77 & 86.4 & 0.68 \\
& Random          & 63.1 & 0.77 & 63.1 & 0.69 \\
& \cellcolor{blue!8}\textbf{Auto-ICL} & \cellcolor{blue!8}\textbf{84.4} & \cellcolor{blue!8}0.75 & \cellcolor{blue!8}\textbf{89.3} & \cellcolor{blue!8}0.70 \\
& \textcolor{gray}{\textit{Oracle}} & \textcolor{gray}{\textit{93.4}} & \textcolor{gray}{\textit{N/A}} & \textcolor{gray}{\textit{91.2}} & \textcolor{gray}{\textit{N/A}} \\
\midrule
\multirow{5}{*}{Countdown}
& Vanilla         & 31.0 & 0.74 & 48.1 & 0.71 \\
& Prefix          & 32.0 & 0.75 & 34.7 & 0.71 \\
& Random          & 17.0 & 0.74 & 21.8 & 0.72 \\
& \cellcolor{blue!8}\textbf{Auto-ICL} & \cellcolor{blue!8}\textbf{42.2} & \cellcolor{blue!8}0.74 & \cellcolor{blue!8}\textbf{49.5} & \cellcolor{blue!8}0.70 \\
& \textcolor{gray}{\textit{Oracle}} & \textcolor{gray}{\textit{49.6}} & \textcolor{gray}{\textit{N/A}} & \textcolor{gray}{\textit{51.3}} & \textcolor{gray}{\textit{N/A}} \\
\bottomrule
\end{tabular}
\label{tab:main_results}
\end{table}

\subsection{Ablation: Shot Count and Generation Length}

\begin{table}[htbp]
\centering
\caption{Comparison of Auto-ICL and Best Static accuracy across varying shot counts. For each task, $p^*$ denotes the optimal global static position index. \textbf{Auto / Best} represents the accuracy (\%) of Auto-ICL and the Best Static placement, respectively, with the superior performance highlighted in bold. Notably, $p^*$ dynamically shifts across different context lengths (e.g., $N \rightarrow N-5$ on GSM8K). Despite these topological shifts, Auto-ICL not only robustly tracks the optimal placement but frequently outperforms the oracle global static baseline (e.g., on Countdown and MATH) by enabling fine-grained, instance-level routing.}
\label{tab:robustness_summary}
\small  
\renewcommand{\arraystretch}{1.1}
\setlength{\tabcolsep}{4pt} % 稍微放大一点列距，因为去掉了竖线
\resizebox{\textwidth}{!}{%
\begin{tabular}{l cccccccc} % 去掉所有竖线，全部改为居中 c
\toprule
\multirow{2}{*}{\textbf{Dataset}} & \multicolumn{2}{c}{\textbf{3-shot}} & \multicolumn{2}{c}{\textbf{4-shot}} & \multicolumn{2}{c}{\textbf{6-shot}} & \multicolumn{2}{c}{\textbf{8-shot}} \\
% 使用带 (lr) 的 cmidrule，让横线左右断开，完美分组！
\cmidrule(lr){2-3} \cmidrule(lr){4-5} \cmidrule(lr){6-7} \cmidrule(lr){8-9}
& $p^*$ & \textbf{Auto / Best (\%)} & $p^*$ & \textbf{Auto / Best (\%)} & $p^*$ & \textbf{Auto / Best (\%)} & $p^*$ & \textbf{Auto / Best (\%)} \\
\midrule
GSM8K     & $N$ & \textbf{62.1}/61.5 & $N$ & \textbf{71.0}/70.8 & $N$ & \textbf{71.3}/69.4 & $N\!-\!5$ & 69.8/\textbf{70.0} \\
Countdown & $0$ & \textbf{40.9}/38.8 & $0$ & \textbf{42.2}/31.0 & $N$ & \textbf{41.3}/40.2 & $N$ & \textbf{42.1}/40.4 \\
MBPP      & $N$ & 46.7/\textbf{48.6} & $N$ & 46.2/\textbf{48.8} & $N$ & \textbf{45.7}/45.2 & $N$ & 45.9/\textbf{46.2} \\
MATH      & $N$ & \textbf{26.3}/25.0 & $N$ & \textbf{28.1}/27.4 & $N$ & \textbf{30.3}/29.0 & $N$ & \textbf{29.8}/29.0 \\
Sudoku    & $0$ & \textbf{83.0}/82.4 & $0$ & \textbf{84.4}/83.0 & $0$ & 83.8/\textbf{84.0} & $0$ & 78.5/\textbf{78.9} \\
\bottomrule
\end{tabular}}
\end{table}

\begin{figure}[htbp]
\centering
\includegraphics[width=\linewidth]{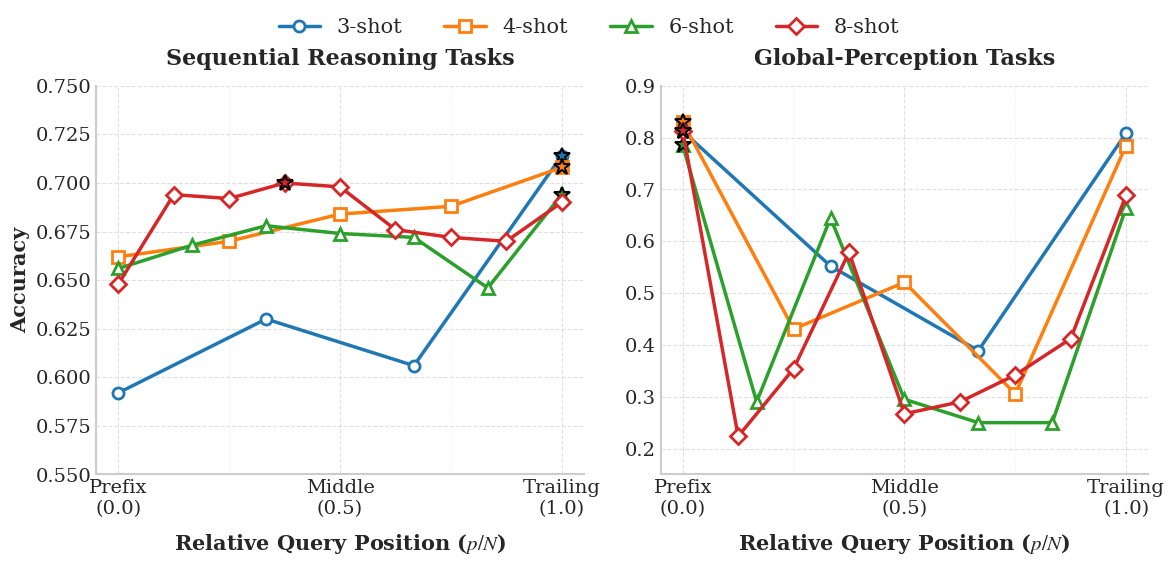}
\caption{Shot ablation across different cognitive task types (accuracy vs. relative query position $p/N$). While Global-Perception tasks strictly maintain prefix-optimality up to 8 shots, Sequential Reasoning tasks exhibit a dominant trailing preference that subtly shifts toward the middle-front under extended contexts (e.g., 8-shot).}
\label{fig:shot_ablation}
\end{figure}

We test whether positional preference remains stable as context grows (Table~\ref{tab:robustness_summary} \& Figure~\ref{fig:shot_ablation}) or generation budget changes (Table~\ref{tab:genlen_ablation}). 

The shot sweep reveals that while broad positional structure persists across different regimes, optimal placements are subject to unpredictable topological shifts. Notably, global-perception tasks (e.g., Sudoku) strictly remain prefix-optimal regardless of context length (up to 8 shots). Interestingly, for sequential reasoning tasks like GSM8K, while trailing placement heavily dominates for standard settings (3 to 6 shots), the optimal position subtly shifts towards the middle-front under extended 8-shot contexts ($p=N-5$). This suggests that when the context becomes excessively long, placing the query at the absolute end might dilute its attention to earlier critical reasoning steps. More importantly, relying on a rigid, fixed prompt template is inherently brittle under such topological shifts. As shown in Table~\ref{tab:robustness_summary}, Auto-ICL successfully navigates this instability. By performing fine-grained, instance-level routing rather than relying on a global dataset-level optimum, Auto-ICL frequently surpasses the theoretical Best Static accuracy (e.g., reaching 42.2\% vs. 31.0\% on 4-shot Countdown, and 30.3\% vs. 29.0\% on 6-shot MATH).

Similarly, we examine the impact of the generation budget on GSM8K (Table~\ref{tab:genlen_ablation}). While increasing the generation budget up to 512 tokens improves absolute accuracy, restricting the budget severely degrades the performance of the globally optimal static placement ($p=4$), dropping to 41.2\% at a 64-token limit. In stark contrast, Auto-ICL dynamically tailors the query placement for each specific instance, enabling it to robustly surpass the global static baseline across all budget constraints. Most notably, Auto-ICL achieves an impressive absolute gain of +6.1\% (47.3\% vs. 41.2\%) under the most restrictive 64-token setting, proving that dynamic spatial routing is especially crucial when generation resources are highly constrained.

\begin{table}[htbp]
\centering
\caption{Generation-length ablation on GSM8K (4-shot). We compare the globally optimal static placement (Best Static) against our dynamic routing approach across different decoding budgets. Remarkably, Auto-ICL consistently outperforms the theoretical global best static baseline across all generation lengths. This highlights the crucial advantage of instance-level spatial routing over rigid formatting, particularly under severe generation constraints (e.g., length 64).}
\label{tab:genlen_ablation}
\small
\renewcommand{\arraystretch}{1.2}
% 【关键修改】：不再手动设置 tabcolsep，使用 tabular* 和 \extracolsep{\fill} 
% 让表格自动撑满文本宽度（\linewidth），左右两端绝对和正文对齐！
\begin{tabular*}{\linewidth}{@{\extracolsep{\fill}}lccc@{}}
\toprule
\textbf{Gen Len} & \textbf{Best Static ($p^*$)} & \textbf{Best Static Acc (\%)} & \textbf{Auto-ICL Acc (\%)} \\
\midrule
64  & $p=4$ & 41.2 & \textbf{47.3} \\
128 & $p=4$ & 68.5 & \textbf{70.3} \\
256 & $p=4$ & 70.8 & \textbf{71.2} \\
512 & $p=4$ & 73.1 & \textbf{75.2} \\
\bottomrule
\end{tabular*}
\end{table}

\subsection{Ablation: Why Single-Step Confidence Fails in dLLMs}
\begin{figure}[htbp]
    \centering
    \includegraphics[width=\textwidth]{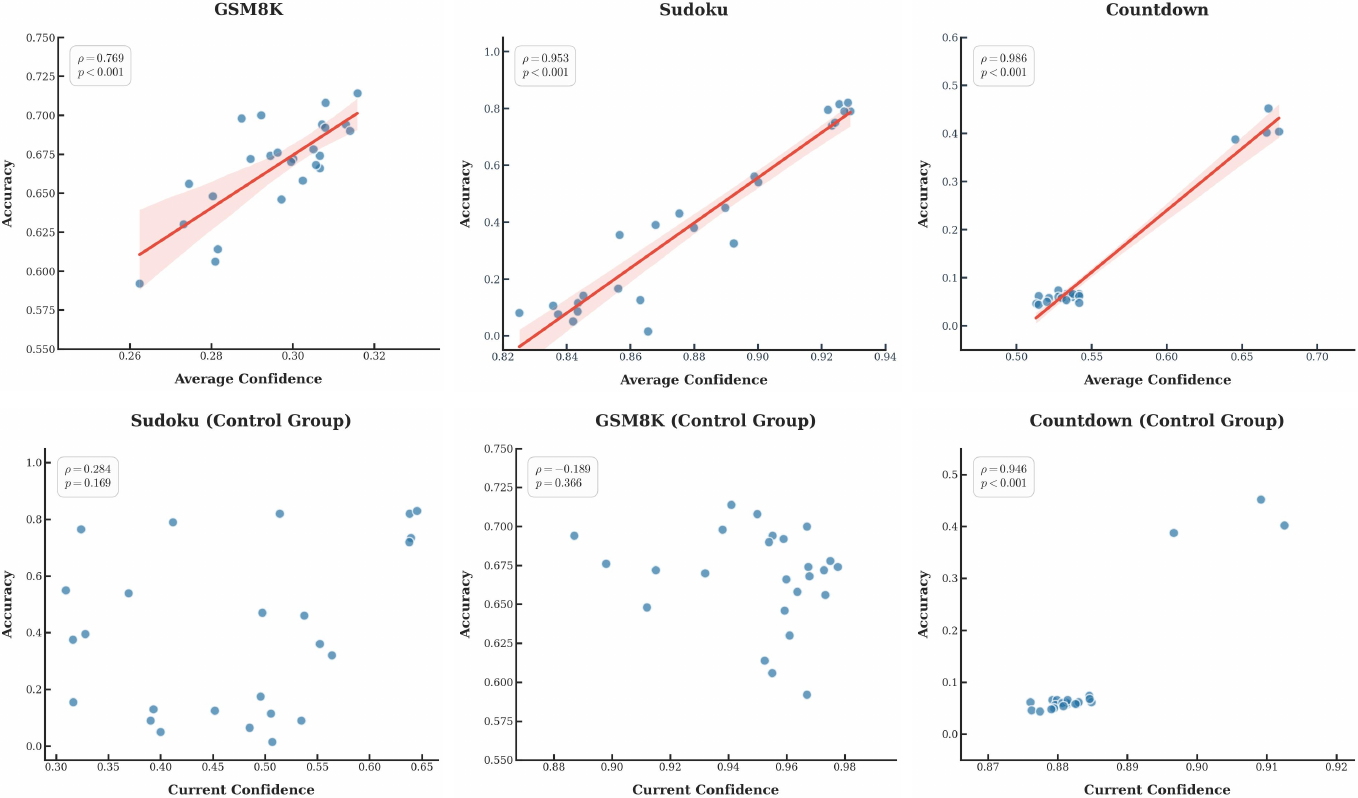}
    \caption{Spearman rank correlation comparison across representative datasets. \textbf{Top:} \textit{Average Confidence ($\overline{C}$)} exhibits a robust positive linear correlation with accuracy across all tasks. \textbf{Bottom:} \textit{Single-step Confidence ($C_{decoded}$, denoted as Current Confidence)} fails to reliably rank positions, demonstrating severe clustering and lack of meaningful correlation.}
    \label{fig:conf_scatter}
\end{figure}

To validate our choice of Average Confidence ($\overline{C}$) as the routing signal, we compare it against the traditional single-step decoding confidence ($C_{decoded}$):
\begin{equation}
C_{decoded} = \frac{1}{|I|} \sum_{i \in I} s_{i}^{(t_i^{decode})}.
\end{equation}
where $t_i^{decode}$ is the decoding step at which token $i$ is first unmasked. As visualized in Figure~\ref{fig:conf_scatter}, we plot the correlation between the metric scores and the downstream accuracy across all spatial positions. 

The bottom row clearly demonstrates that $C_{decoded}$ completely fails in dLLMs. The scores severely cluster near $1.0$ regardless of the actual generation accuracy, indicating that looking only at the final decoding step masks the underlying reasoning instability. In stark contrast, our proposed $\overline{C}$ (top row), which aggregates the activation probabilities across the entire temporal decoding trajectory, exhibits a highly robust and positive linear correlation with task accuracy. This confirms that sequence-level generation stability in bidirectional dLLMs must be evaluated temporally, proving the indispensability of $\overline{C}$ for our Auto-ICL routing strategy.

% --- Section 8: CONCLUSION ---
\section{Conclusion and Future Work}
We uncover and mitigate the severe positional bias of ICL in Diffusion LLMs. Through rigorous empirical decoupling, we prove that query position is a first-order variable. Mechanistically, this bias stems from a spatial Recency Effect and a temporal shift in Decoding Trajectories. By utilizing Average Confidence ($\overline{C}$) to holistically measure decoding stability, our proposed Auto-ICL routing strategy successfully identifies optimal query topologies without training or labels. 

While Auto-ICL significantly bridges the gap to oracle performance, exhaustively evaluating all candidate positions introduces computational overhead ($N+1$ forward passes per query). Future work will explore lightweight early-exit predictors and Beam Search heuristics to drastically reduce this inference cost, further enabling efficient, dynamically configured ICL in bidirectional contexts.

% --- Bibliography ---
\bibliographystyle{splncs04}
\bibliography{references}

\end{document}